\documentclass[letterpaper]{article} 
\usepackage{aaai24}  
\usepackage{times}  
\usepackage{helvet}  
\usepackage{courier}  
\usepackage[hyphens]{url}  
\usepackage{graphicx} 
\usepackage{natbib}  
\usepackage{caption} 
\usepackage{algorithm}
\usepackage{algorithmic}

\usepackage{newfloat}
\usepackage{listings}
\DeclareCaptionStyle{ruled}{labelfont=normalfont,labelsep=colon,strut=off} 
\floatstyle{ruled}
\newfloat{listing}{tb}{lst}{}
\floatname{listing}{Listing}
\usepackage{booktabs} 
\usepackage{multirow}
\usepackage{amssymb}
\usepackage{pifont}
\usepackage{arydshln}
\usepackage{amsmath}
\usepackage{adjustbox}
\usepackage{subfigure}

\title{Direction-aware Video Demoiréing with Temporal-guided Bilateral Learning}
\nocopyright
\author{
        Shuning Xu\textsuperscript{\rm 1}\equalcontrib,	
        Binbin Song\textsuperscript{\rm 1}\equalcontrib, 
	Xiangyu Chen\textsuperscript{\rm 1,2}, and 
	Jiantao Zhou\textsuperscript{\rm 1}\thanks{Corresponding Author.} \\
}

\affiliations{
    \textsuperscript{\rm 1}State Key Laboratory of Internet of Things for Smart City, University of Macau\\
    \textsuperscript{\rm 2} Shenzhen Institutes of Advanced Technology, Chinese Academy of Sciences \\

    \{yc07425, yb97426, jtzhou\}@um.edu.mo, chxy95@gmail.com
}

\usepackage{bibentry}

\begin{document}

\maketitle

\begin{abstract}
Moiré patterns occur when capturing images or videos on screens, severely degrading the quality of the captured images or videos. Despite the recent progresses, existing video demoiréing methods neglect the physical characteristics and formation process of moiré patterns, significantly limiting the effectiveness of video recovery. This paper presents a unified framework, DTNet, a direction-aware and temporal-guided bilateral learning network for video demoiréing. DTNet effectively incorporates the process of moiré pattern removal, alignment, color correction, and detail refinement. Our proposed DTNet comprises two primary stages: Frame-level Direction-aware Demoiréing and Alignment (FDDA) and Tone and Detail Refinement (TDR). In FDDA, we employ multiple directional DCT modes to perform the moiré pattern removal process in the frequency domain, effectively detecting the prominent moiré edges. Then, the coarse and fine-grained alignment is applied on the demoiréd features for facilitating the utilization of neighboring information. In TDR, we propose a temporal-guided bilateral learning pipeline to mitigate the degradation of color and details caused by the moiré patterns while preserving the restored frequency information in FDDA. Guided by the aligned temporal features from FDDA, the affine transformations for the recovery of the ultimate clean frames are learned in TDR. Extensive experiments demonstrate that our video demoiréing method outperforms state-of-the-art approaches by 2.3 dB in PSNR, and also delivers a superior visual experience. Our code is available at https://github.com/rebeccaeexu/DTNet.

\end{abstract}

\section{Introduction}
Moiré patterns appear when two similar repetitive patterns interfere with each other, a phenomenon commonly observed during image capture on screens. The occurance of moiré patterns can be intricate and multifaceted, leading to an unsatisfactory visual experience. Eliminating moiré patterns can be arduous owing to their ambiguous shapes, diverse colors, and variable frequencies.

\begin{figure}[!ht]
\centering
\includegraphics[width=1\linewidth]{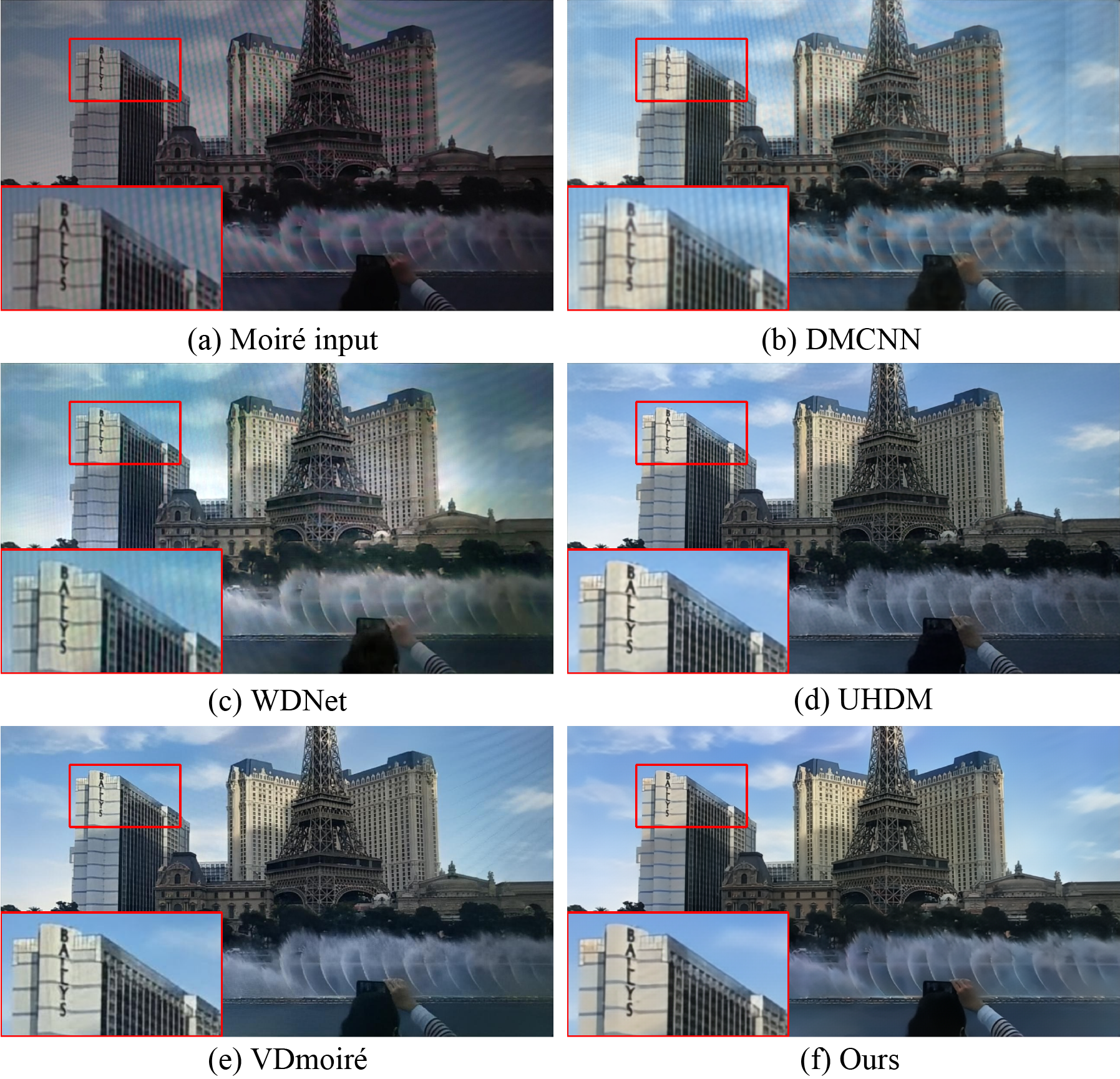}
\caption{A visual comparison with the existing
methods for video demoiréing. (a) Moiré input, (b-d) Outputs of state-of-the-art image demoiréing methods, (e) Output of the video demoiréing method VDmoiré, (f) Our result, capable of removing the moiré patterns, while restoring color deviations and fine details. (Zoom in for a better view.)}
\label{fig:introduction}
\end{figure}

Image demoiréing has received increasing attention from the research community.
DMCNN ~\cite{sun2018moire} proposes the first deep end-to-end network that uses a multi-scale architecture to remove moiré patterns of various sizes. 
MBCNN ~\cite{zheng2020image}, WDNet ~\cite{liu2020wavelet} and FHDe2Net ~\cite{he2020fhde} compensate the fine detail distortions in demoiréing by exploiting signal properties in the frequency domain. 
ESDNet ~\cite{yu2022towards} develops a lightweight model for high-resolution image demoiréing. 
The aforementioned models are designed for removing moiré patterns from a single image, and there are relatively fewer algorithms targeting at removing moiré patterns from videos. If we directly apply image demoiréing methods to remove moiré patterns from videos, it may result in poor temporal consistency. This limitation arises from the inability of image demoiréing methods to make use of information from neighboring frames for restoring moiré patterns in the current frame ~\cite{dai2022video}. Hence, the removal of moiré patterns from videos necessitates the development of algorithms tailored to address this issue.
VDmoiré ~\cite{dai2022video} builds the first video demoiréing dataset captured by hand-held cameras and designs a baseline video demoiring model that can effectively leverage nearby frames with the relation-based consistency regularization. 
FPANet ~\cite{oh2023fpanet} removes moiré patterns and recovers the original color in the frequency domain using amplitude and phase.
However, existing restoration methods lack modules specifically designed for the unique characteristics of moiré patterns, and they neglect the formation process of moiré patterns in videos. As a result, unsatisfactory video demoiréing performance has been observed in these methods. 

This work presents a direction-aware and temporal-guided bilateral learning video demoiréing network (DTNet), a unified architecture that combines moiré pattern removal, alignment, color correction, and detail refinement. Our proposed DTNet is structured into two key stages: Frame-Level Direction-Aware Demoiréing and Alignment (FDDA), and Tone and Detail Refinement (TDR).
Within FDDA, a direction-aware demoiréing module with eight predefined directions is designed to eliminate moiré patterns from each frame of the input moiré video, identifing the prominent moiré edges within each block in the Discrete Cosine Transform (DCT) domain.
Furthermore, we introduce a learnable band reject filter (LBRF) in the direction-aware demoiré module to attenuate the specific frequencies of moiré patterns.
Also, the coarse-to-fine alignment is applied on the demoiréd features for facilitating better utilization of the neighboring information.
In TDR, a temporal-guided bilateral algorithm is formed to address the color deviation and restore the texture of the original content. We propose to learn a 3D bilateral grid, storing affine coefficients and biases, based on pixel position and intensity for spatially variant color restoration while respecting the edges. For better utilization of temporal information, guidance maps are extracted from adjacent frames, furnishing valuable temporal cues for effective color and detail refinement.
Fig.~\ref{fig:introduction} presents a visual comparison with existing methods on a video with moiré patterns. Notably, our proposed DTNet effectively eliminates moiré patterns, simultaneously restoring color deviations and fine details.

In summary, our contributions are listed as follows:
\begin{itemize}
\item[$\bullet$] We design a direction-aware and temporal-guided bilateral learning network (DTNet) for video demoiréing. DTNet effectively incorporates moiré pattern removal, alignment, color correction, and detail refinement.
\item[$\bullet$] With the consideration of the unique characteristics of moiré patterns, we propose a directional-aware demoiréing module for moiré pattern removal.
\item[$\bullet$] For the color restoration and detail refinement of the final output, we propose a novel temporal-guided bilateral learning strategy, promoting the spatially variant color recovery while respecting the edges of latent clean images.
\item[$\bullet$] Extensive experiments are conducted on the video demoiréing dataset. Compared with state-of-the-art methods, our DTNet achieves a dominant performance gain in both qualitative and quantitative evaluations.
\end{itemize}

\section{Related Works}
\subsection{Image and Video Demoiréing}
Moiré arises from the interference of two patterns with similar frequencies and often occurs when capturing screen images, resulting in significant degradation of image quality.
To remove the moiré patterns from the original images, many end-to-end image demoiréing solutions have been proposed ~\cite{liu2018demoir, cheng2019multi, liu2020self, liu2020mmdm, wang2021image, niu2023progressive, wang2023coarse, zhang2023real}. 
DMCNN~\cite{sun2018moire} presents the first real-world dataset for image demoiréing and a multi-resolution convolutional neural network. 
MopNet ~\cite{he2019mop} designs a multi-scale aggregated, edge-guided, and pattern attribute-aware network for moiré pattern removal.
In addition to the elaborate designs on the spatial domain, several studies utilize frequency domain learning for image demoiréing ~\cite{liu2020wavelet, he2020fhde, zheng2020image}.
WDNet~\cite{liu2020wavelet} decomposes the input image with moiré patterns into different frequency bands using a wavelet transform and designs a dual-branch network for restoring the close-range and far-range information.
MBCNN~\cite{zheng2020image} utilizes learnable bandpass filters to acquire a frequency prior to separate moiré patterns from normal image texture.
In addition, there are also networks specifically designed for video demoiréing ~\cite{dai2022video, oh2023fpanet}.
VDMoiré~\cite{dai2022video} presents a simple video demoiréing model that utilizes multiple video frames. It also employs a novel relation-based consistency loss, which enhances the temporal consistency of videos.
FPANet~\cite{oh2023fpanet} learns filters in both frequency and spatial domains, enhancing the restoration quality by eliminating moiré patterns of different sizes.
Note that existing demoiréing methods have not been designed with modules specific to the directional characteristics of moiré patterns for restoration. This issue will be explicitly addressed in our proposed DTNet.

\begin{figure*}[!ht]
\centering
\includegraphics[width=0.97\linewidth]{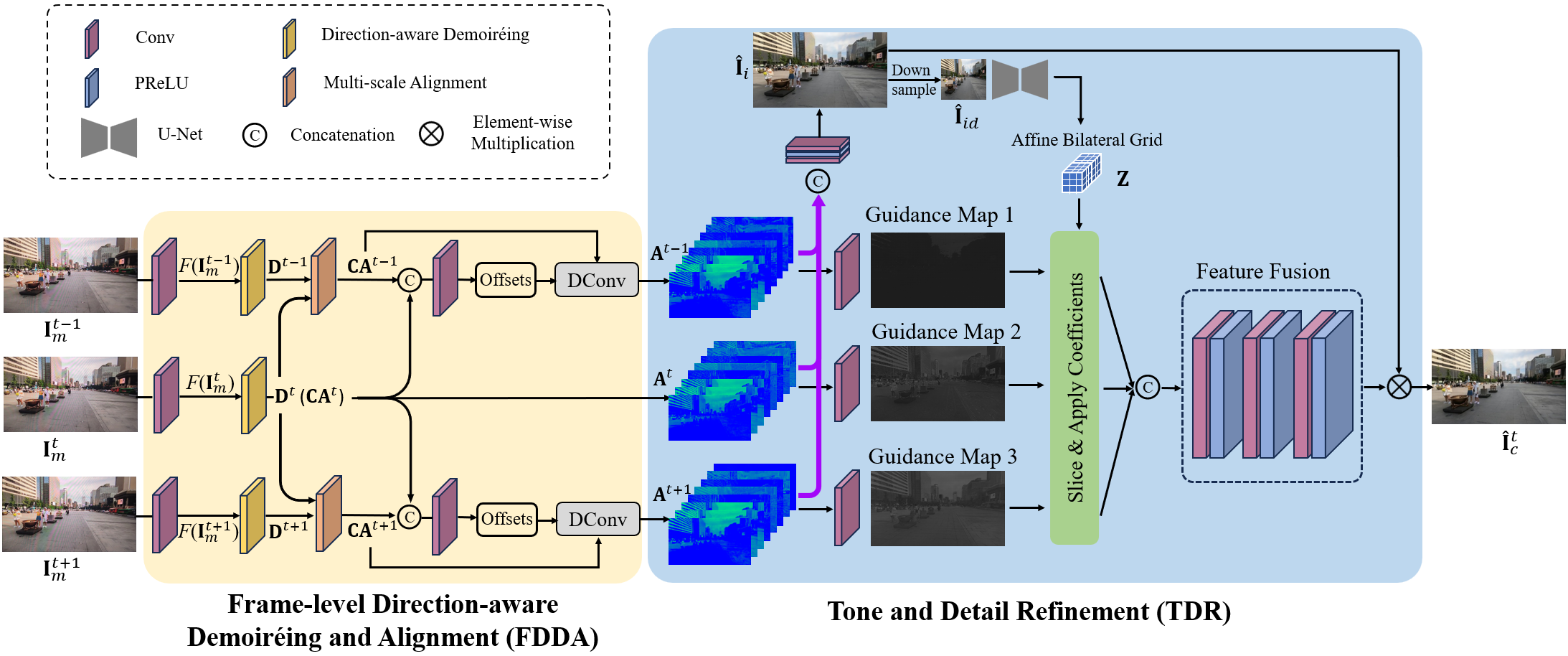}
\caption{The overview of our proposed DTNet for video demoiréing. The whole framework consists of two stages: (\romannumeral1) Frame-level Direction-aware Demoiréing and Alignment (FDDA), and (\romannumeral2) Tone and Detail Refinement (TDR).}
\label{fig:architecture}
\end{figure*}

\subsection{Bilateral Filtering}
The bilateral filter is a non-linear smoothing filter that preserves image edges and reduces noise ~\cite{tomasi1998bilateral, banterle2012low}. It has garnered significant attention for its ability to accelerate the edge-aware manipulation of images in the bilateral space ~\cite{barron2016fast, chen2007real, chen2017fast, pham2005separable, gavaskar2018fast, yang2012recursive, zhang2008adaptive}. Recently, various studies focus on the application of bilateral filters for image enhancement ~\cite{gharbi2017deep, xia2020joint, zheng2021ultrahdr, xu2021deep, ren2020single}. 
Zhu \textit{et al.} ~\cite{zheng2021ultradehazing} reconstructs bilateral coefficients on a reduced resolution of the input and generates high-quality feature maps under the guidance of full-resolution features. 
Xu \textit{et al.}~\cite{xu2021bilateral} proposes an edge-aware cost volume upsampling module, regressing the disparity map at a high resolution to keep the high accuracy, while maintaining high efficiency.
As mentioned above, the bilateral filter has been previously utilized mainly for image restoration purposes, while no applications have yet been found in video restoration problems. 
In this paper, we devise a novel technique called temporal-guided bilateral learning by extending the use of bilateral filters from individual frames to multiple frames. The temporal-guided bilateral module will be shown effective in restoring both the color and edges in videos.

\section{Proposed Method}
For the image with moiré patterns $\mathbf{I}_m$, the removal process of moiré patterns $\mathbf{M}$ ~\cite{zheng2020image} can be modeled as:

\begin{equation}
\small
    \mathbf{\hat{I}}_c = \mathcal{T}(\mathbf{I}_m - \mathbf{M}),
   \label{equ:imagedemoire}
\end{equation}
where $\mathbf{\hat{I}}_c$ denotes the estimation of the clean image displayed on the screen. $\mathcal{T}$ refers to the correction process of the color shift, caused by the ambient light and the screen itself.

When it comes to removing moiré patterns from videos, the input consists of multiple frames with moiré patterns. Here, we take the example of three consecutive frames at timestamp \{$t-1$, $t$, $t+1$\} as input. The process of removing moiré patterns from videos can be inferred as:
\begin{equation}
\small
    \begin{split}
        \mathbf{A}^{t+i} = \Lambda((&\mathbf{I}^{t+i}_m - \mathbf{M}^{t+i}), (\mathbf{I}^t_m-\mathbf{M}^t)), \\
        \quad &i\in\{-1, 0, 1\},
    \end{split}
    \label{equ:videodemoire_1}
\end{equation}
\begin{equation}
\small
    \mathbf{\hat{I}}^t_c = \Gamma(\mathbf{A}^{t-1}, \mathbf{A}^{t}, \mathbf{A}^{t+1}),  
    \label{equ:videodemoire_2}
\end{equation}
where $\mathbf{I}^{t+i}_m$ and $\mathbf{M}^{t+i}$ respectively denote the frames of $\mathbf{I}_m$ and $\mathbf{M}$ at the specific timestamp  $t+i$. 
$\Lambda$ is the alignment process, where the consecutive demoiréd frames are aligned to the reference one at timestamp $t$. $\mathbf{A}^{t+i}$ denotes the aligned deep features of the frame at timestamp $t+i$. 
$\Gamma$ denotes the color refinement process on the aligned features. In this manner, the task of video demoiréing can be separated into two steps: \romannumeral1) moiré pattern removal and alignment, and \romannumeral2) color restoration and detail refinement.

In this work, we propose a direction-aware and temporal-guided bilateral learning network (DTNet) for video demoiréing, where a single network incorporates moiré pattern removal, alignment, color correction, and detail refinement. Abiding by ~\eqref{equ:videodemoire_1} and~\eqref{equ:videodemoire_2}, we categorize the implementation of the aforementioned functions into FDDA and TDR. As demonstrated in Fig.~\ref{fig:architecture}, the consecutive input frames with moiré patterns $\{\mathbf{I}^{t-1}_m, \mathbf{I}^{t}_m, \mathbf{I}^{t+1}_m\}$ undergo processing in the direction-aware demoiréing module and the coarse-to-fine alignment process within FDDA. In TDR, color and texture features are adaptively fused in the dual-path network to generate the final output image $\mathbf{\hat{I}}^t_c$.

\subsection{Design of FDDA}
Given three consecutive frames \{$\mathbf{I}^{t-1}_m$, $\mathbf{I}^{t}_m$, $\mathbf{I}^{t+1}_m$\}, we propose FDDA to remove moiré patterns in each frame and then align the features of neighboring frames, i.e., $\mathbf{I}^{t-1}_m$ and $\mathbf{I}^{t+1}_m$, to the reference frame $\mathbf{I}^{t}_m$.

Considering the formation process of moiré patterns, various methods have been proposed to separate moiré patterns and image content using frequency domain analysis, which is considered as a preferable approach. This is beneficial as image signal and moiré patterns are usually more separable from a frequency perspective.  
Therefore, to obtain the aligned features $\{\mathbf{A}^{t-1}, \mathbf{A}^{t}, \mathbf{A}^{t+1}\}$, we firstly extract the shallow feature, noted as $F(\mathbf{I}^{t+i}_m)$, from $\mathbf{I}^{t+i}_m$. Then Block-DCT is adopted to handle the moiré pattern removal on $F(\mathbf{I}^{t+i}_m)$ in the frequency domain. Here, we denote $\mathbf{S}^{t+i}_{j}$ and  $\mathbf{R}^{t+i}_{j}$ respectively as the frequency spectrum of $F(\mathbf{I}^{t+i}_m)$ and $F(\mathbf{M}^{t+i})$ at the frequency $j$, where $F(\mathbf{M}^{t+i})$ is the shallow feature of $\mathbf{M}^{t+i}$. Thus, the DCT transformations of $F(\mathbf{I}^{t+i}_m)$ and $F(\mathbf{M}^{t+i})$ can be computed by:
\begin{equation}
\small
\begin{split}
	DCT(F(\mathbf{I}^{t+i}_m)) \!=\! \sum\nolimits_{j}\mathbf{S}^{t+i}_{j}, \quad DCT(F(\mathbf{M}^{t+i})) \!= \!\sum\nolimits_{j}\mathbf{R}^{t+i}_{j}.
	\label{eq:dct}
\end{split}
\end{equation}

Owning to the linear property of Block-DCT/IDCT, the demoiréd features $\mathbf{D}^{t+i}$ can be acquired by:
\begin{equation}
\begin{small}
\begin{aligned}
    \mathbf{D}^{t+i} =& F(\mathbf{I}_m^{t+i}) - F(\mathbf{M}^{t+i}) \\
    =& IDCT(DCT(F(\mathbf{I}_m^{t+i}))) \!- \!IDCT(DCT(F(\mathbf{M}_m^{t+i}))) \\
         = & IDCT(\sum\nolimits_{j} \mathbf{S}^{t+i}_{j} \!- \! \sum\nolimits_{j}\mathbf{R}^{t+i}_{j}),
\end{aligned}
\label{eq:moire_dct}	
\end{small}
\end{equation}
where $IDCT$ indicates the inverse function of block-DCT.

\begin{figure}[htbp]
\centering
\subfigure[Eight directional DCT modes.]{
\includegraphics[width=0.75\linewidth]{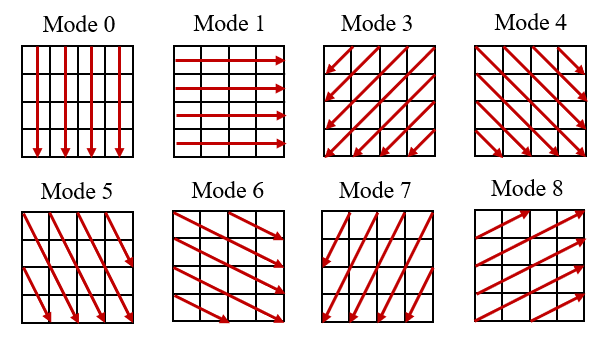} 
}
\subfigure[Direction-aware Demoiréing module.]{
\includegraphics[width=0.9\linewidth]{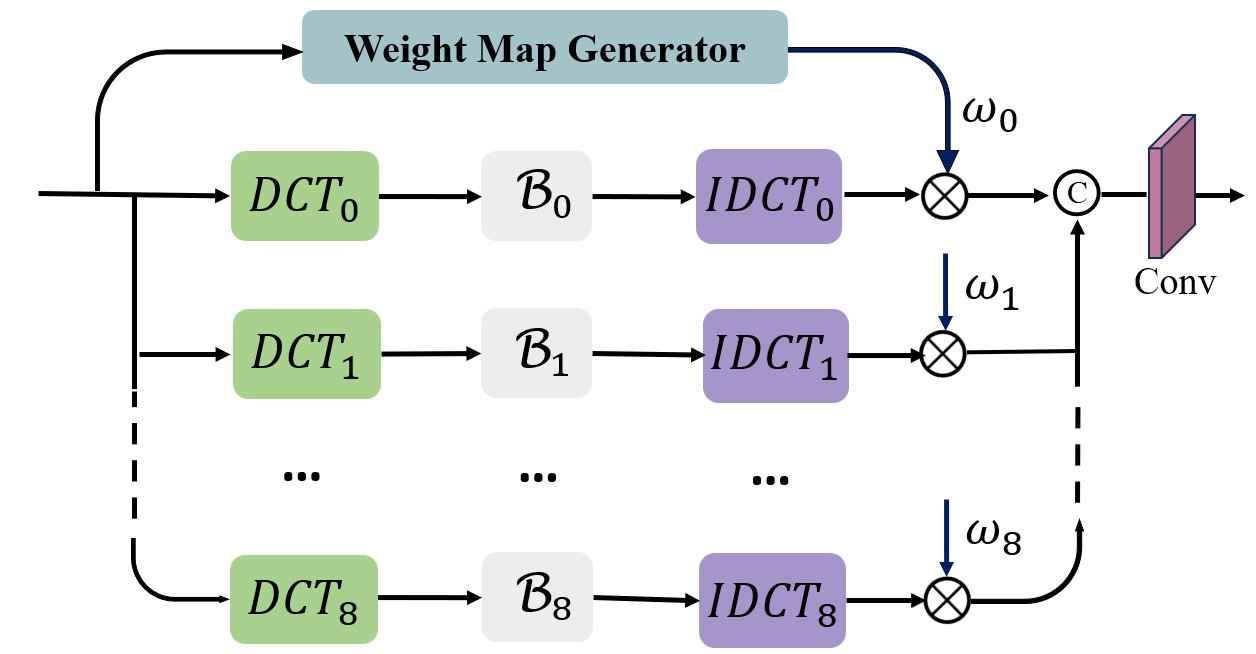} 
}
\DeclareGraphicsExtensions.
\caption{(a) Eight directional DCT modes. (b) The structure of the proposed Direction-aware Demoiréing module.}
\label{fig:dd}
\end{figure}

We assume that the frequency spectrum of moire patterns tends to be consistent within a small patch. In signal processing, a band reject filter (BRF) is a filter that passes most frequencies unaltered, but attenuates those in a specific range to very low levels. Hence, the application of a BRF becomes a viable means for effective moiré pattern removal. Since different patches necessitate the removal of distinct frequencies, determining a frequency prior for each moiré image patch becomes time-consuming. Consequently, we construct a learnable BRF (LBRF), denoted as $\mathcal{B}(\cdot)$, with several convolutional layers to attenuate the specific frequencies of moiré patterns while preserving the orignal image contents. The removal of moiré patterns in the frequency domain is defined as:
\begin{small}
\begin{equation}
	\mathcal{B}(\sum\nolimits_{j}\mathbf{S}^{t+i}_{j}) = \sum\nolimits_{j}\mathbf{S}^{t+i}_{j} - \sum\nolimits_{j}\mathbf{R}^{t+i}_{j}.
	\label{eq:lbrf}
\end{equation}
\end{small}
Substituting ~\eqref{eq:lbrf} into ~\eqref{eq:moire_dct}, the moiré pattern removal process in the deep feature space can be rewritten as:
\begin{equation}
\small
	\begin{split}
	\mathbf{D}^{t+i} & = IDCT(\mathcal{B}(\sum\nolimits_{j}\mathbf{S}^{t+i}_{j})) \\
	& = IDCT(\mathcal{B}(DCT(F(\mathbf{I}^{t+i}_m)))).
	\end{split}
	\label{eq:brf}
\end{equation}
Inherently, the human eye possesses high sensitivity in detecting edges within an image. For conventional DCT, it is executed through two separate 1-D transformations, applied along the vertical and horizontal directions. Nevertheless, the moiré patterns exhibit irregular shapes characterized by distinct directions beyond the vertical or horizontal ones. Inspired by the directional DCT works~\cite{zeng2008directional}, we introduce eight directional modes in DCT and IDCT, which are defined similarly to those used in the H.264 standard (the dc mode, Mode 2, is not counted), as depicted in Fig.~\ref{fig:dd} (a). In Zeng's work, they adopt a brute-force method in choosing the most appropriate directional mode, which is time-consuming and vulnerable to a wrong decision.
Therefore, we propose a new direction-aware demoiréing (DD) module that incorporates eight pre-defined directions within individual branches to effectively detect prominent moiré edges in the DCT domain, as demonstrated in Fig.~\ref{fig:dd} (b).

In FDDA, with the input features of $\mathbf{I}^{t+i}_m$, DD aims at acquiring the demoiréd features $\mathbf{D}^{t+i}$, $i\in \{{-}1, 0, {+}1\}$.
To distinguish the importance of demoiréd features in different directions and facilitate feature aggregation, we generate weight maps $\omega$ with two convolutional layers and a sigmoid function and then split them into eight attention weights $\{\omega_0, \omega_1, \omega_3, ... , \omega_8\}$ for each branch. 
Followed by the multiplication of spatial content features with the corresponding weight map in an element-wise manner, we concatenate the results and adopt a 3$\times$3 convolution to generate the output features, effectively suppressing moiré patterns.

After we perform the aforementioned pre-demoiré process for each frame, shown in the left part of FDDA in Fig.~\ref{fig:architecture}, we adopt the multi-scale alignment operation in EDVR~\cite{wang2019edvr} to acquire coarsely aligned features, denoted as $\mathbf{CA}^{t+i},i\in \{{-}1, 0, {+}1\}$. To further refine the coarsely aligned features, we predict the learnable offset and utilize deformable convolution (DConv) to align each neighboring frame to the reference frame at the feature level, generating the aligned feature $\mathbf{A}^{t+i},i\in \{{-}1, 0, {+}1\}$, at each position. Followed by the aforementioned steps, we can obtain the moiré-suppressed and  aligned features $\{\mathbf{A}^{t-1}, \mathbf{A}^{t}, \mathbf{A}^{t+1}\}$.

\begin{table*}[htbp]
    \begin{center}
    \footnotesize
    \caption{Quantitative comparison with the state-of-the-art image (or video) demoiréing approaches. The best results are highlighted with \textbf{bold}. The second-best results are highlighted with \underline{underline}.}
    \label{table:quantative-sota}
    \resizebox{\textwidth}{!}{
        \begin{tabular}{lccc|ccc|ccc|ccc}
        \toprule
        \multirow{2}{*}{Method} & \multicolumn{3}{c|}{TCL-V1} & \multicolumn{3}{c|}{TCL-V2} & \multicolumn{3}{c|}{iPhone-V1} & \multicolumn{3}{c}{iPhone-V2} \\ \cmidrule{2-13}
         &  PSNR$\uparrow$ & SSIM$\uparrow$ & LPIPS$\downarrow$ & PSNR$\uparrow$ & SSIM$\uparrow$ & LPIPS$\downarrow$ &  PSNR$\uparrow$ & SSIM$\uparrow$ & LPIPS$\downarrow$&  PSNR$\uparrow$ & SSIM$\uparrow$ & LPIPS$\downarrow$\\\midrule
         
        DMCNN  & 20.321 & 0.703 & 0.321 & 20.707 & 0.793 & 0.385 & 21.967 & 0.712  & 0.280 & 21.816 & 0.749 & 0.496 \\
        WDNet & 19.650 & 0.726 & 0.289 & 20.334 & 0.847 & 0.288 & 19.818& 0.722 & 0.300 & 20.613 & 0.832 & 0.297 \\
        ESDNet & \underline{22.026} & \underline{0.734} & \underline{0.199} & \underline{24.896} & \underline{0.874} & \underline{0.165} & \underline{22.537} & \underline{0.731} & \underline{0.218} & 25.064 & 0.853 & 0.165 \\
        VDmoiré & 21.725 & 0.733 & 0.202 & 23.460 & 0.857 & 0.163  & 21.990 & 0.707 & 0.221  & \underline{25.230} & \bf{0.860} & \underline{0.157}\\  
        Ours & \bf{24.119} & \bf{0.801} & \bf{0.163} & \bf{26.153} & \bf{0.877} & \bf{0.128} & \bf{24.821} & \bf{0.794} & \bf{0.172} & \bf{26.503} & \underline{0.854} & \bf{0.149}\\
        \cmidrule{2-13}
         & (+2.093) & (+0.067) & (-0.036) & (+1.257) & (+0.003) & (-0.037) & (+2.284) &  (+0.063) & (-0.046) & (+1.273) & (-0.006) & (-0.008)\\
        \bottomrule
        
    \end{tabular}}
\end{center}
\end{table*}

\subsection{Design of TDR}
TDR is designed to restore the ultimate clean result $\mathbf{\hat{I}}^t_c$ from the aligned features $\{\mathbf{A}^{t-1}, \mathbf{A}^{t}, \mathbf{A}^{t+1}\}$. In TDR, we aim to refine the tone and color details degraded by the moiré patterns while preserving the edges of the latent clean image. To this end, inspired by the bilateral filtering ~\cite{chen2007real, gharbi2017deep, xia2020joint}, we propose a temporal-guided bilateral learning framework for the video restoration in TDR.

The specific design of TDR is depicted in the right portion of Fig.~\ref{fig:architecture}. For the color refinement in the RGB space, we first generate a full-size intermediate result $\hat{\mathbf{I}}_i$ from the aligned features $\{\mathbf{A}^{t-1}, \mathbf{A}^{t}, \mathbf{A}^{t+1}\}$ with a combination of two convolutional layers and a PReLU layer. The color refinement can be achieved by fitting local affine transformations ~\cite{luan2017deep, xia2020joint} from the intermediate result $\hat{\mathbf{I}}_i$ to the ultimate output $\mathbf{\hat{I}}^t_c$. To this end, we propose to learn a 3D bilateral grid $\mathbf{Z}$ to store the affine coefficients and biases. Then, to exploit the temporal information for the guidance of the color refinement, we construct guidance maps from $\{\mathbf{A}^{t-1}, \mathbf{A}^{t}, \mathbf{A}^{t+1}\}$ to query the affine parameters in $\mathbf{Z}$ for the mapping from $\mathbf{\hat{I}}_i$ to $\mathbf{\hat{I}}^t_c$.

The upper part of TDR demonstrates the diagram of the generation of $\mathbf{Z}$. To save the memory consumption and accelerate the computation, we learn $\mathbf{Z}$ from the downsampled $\mathbf{\hat{I}}_i$, which is denoted as $\mathbf{\hat{I}}_{id}$ with the size $256 \times 256 \times 3$, rather than the original full-resolution version. The bilateral grid $\mathbf{Z}$ is learned through a U-Net and reorganized into a 3D array with the size $16 \times 16 \times 16$. In the bottom part of TDR, to exploit the information of adjacent frames to guide the color and detail restoration of the reference frame, we use a single convolutional layer to generate three guidance maps $\{\mathbf{G}^{t-1}, \mathbf{G}^{t}, \mathbf{G}^{t+1}\}$ from the aligned features $\{\mathbf{A}^{t-1}, \mathbf{A}^{t}, \mathbf{A}^{t+1}\}$. For a specific pixel with the coordinate $(x, y)$ in the single-channel temporal guidance map $\mathbf{G}^{t+i}$, we slice $\mathbf{Z}$ and look up the affine transformation coefficients $\mathbf{W}^{t+i}$ and biases $\mathbf{B}^{t+i}$ by:
\begin{equation}
\small
    \mathbf{W}^{t+i}, \mathbf{B}^{t+i} = \mathbf{Z}_{\lfloor\frac{x}{h} \times 16\rfloor, \lfloor\frac{y}{w} \times 16\rfloor, \lfloor \mathbf{G}^{t+i}_{x,y} \times 16\rfloor},\quad i\in\{-1, 0, 1\},
\end{equation}
where $w$ and $h$ respectively denote the weight and height of $G^{t+i}$, and $\lfloor\cdot \rfloor$ means the rounding operator. By taking into account the query pixel intensities, we preserve the background edges recovered in FDDA. Lastly, the ultimate result $\mathbf{\hat{I}}^t_c$ is generated by:
\begin{equation}
\small
\mathbf{\hat{I}}^t_c \! = \! Conv([\mathbf{W}^{t-1}, \mathbf{W}^{t}, \mathbf{W}^{t+1}]) \otimes \mathbf{I}_{i}  + Conv([\mathbf{B}^{t-1}, \mathbf{B}^{t}, \mathbf{B}^{t+1}]), 
\end{equation}
where $\otimes$ represents the Hadamard product and $[\cdot]$ means the concatenation operation.

\subsection{Training Objectives}
We train our framework in an end-to-end manner, and the overall training objective can be expressed as:
\begin{equation}
\small
\label{eq:loss}
\mathcal{L} = \|\hat{\mathbf{I}}^t_c - \mathbf{I}^t_{c}\|_1 + \| \Phi_l(\hat{\mathbf{I}}^t_c) - \Phi_l(\mathbf{I}^t_{c}) \|_1,
\end{equation}
where $\mathbf{I}^t_{c}$ is the ground-truth moiré-free image of the frame $t$.
We employ $L_1$ loss in conjunction with perceptual loss~\cite{johnson2016perceptual}, which can reflect the human visual system's perception of image quality. $\Phi_l(\cdot)$ denotes a set of VGG-16 layers. 

\section{Experiments}

\subsection{Experimental Setup}

\subsubsection{Dataset}
We evaluate the effectiveness of our proposed methods using the VDmoiré dataset~\cite{dai2022video}. This dataset consists of 290 clean source videos and the corresponding moiréd videos. The source videos, with a resolution of 720p (1080$\times$720), are displayed on either the MacBook Pro display or the Huipu v270 display. A hand-held camera (either iPhoneXR or TCL20 pro camera) captures the screen, resulting in recorded frames with moiré patterns. To minimize the impact of misaligned frame correspondences, two methods are employed: estimating the homography matrix (referred as V1) and applying optical flow (referred as V2) to align the frames. To compare our proposed methods with state-of-the-art approaches in diverse settings, we carry out our experiments on four different dataset settings: TCL-V1, TCL-V2, iPhone-V1, and iPhone-V2.

\subsubsection{Training Details}
The video demoiréing network utilizes three consecutive frames as input to produce one restored clean image. We adopt the AdamW optimizer with $\beta_1$ = 0.9 and $\beta_2$ = 0.999 to train the model. The learning rate is initialized as 4$\times10^{-4}$. We apply the cyclic cosine annealing learning rate schedule~\cite{loshchilov2016sgdr}, which allows partial warm restart optimization, generally improving the convergence rate in gradient-based optimization. In total, we train our model with batch size 16 on four NVIDIA Tesla A100 GPUs.

\subsection{Frame-Level Comparison}
We compare our approach with several demoiréing methods: DMCNN ~\cite{sun2018moire}, WDNet ~\cite{liu2020wavelet}, ESDNet ~\cite{yu2022towards}, and VDmoiré ~\cite{dai2022video}.
It is important to note that both VDMoiré and our method utilize multi-frame inputs, whereas other methods can only accept single-frame images as input.

\begin{figure*}[!ht]
\centering
\includegraphics[width=1\linewidth]{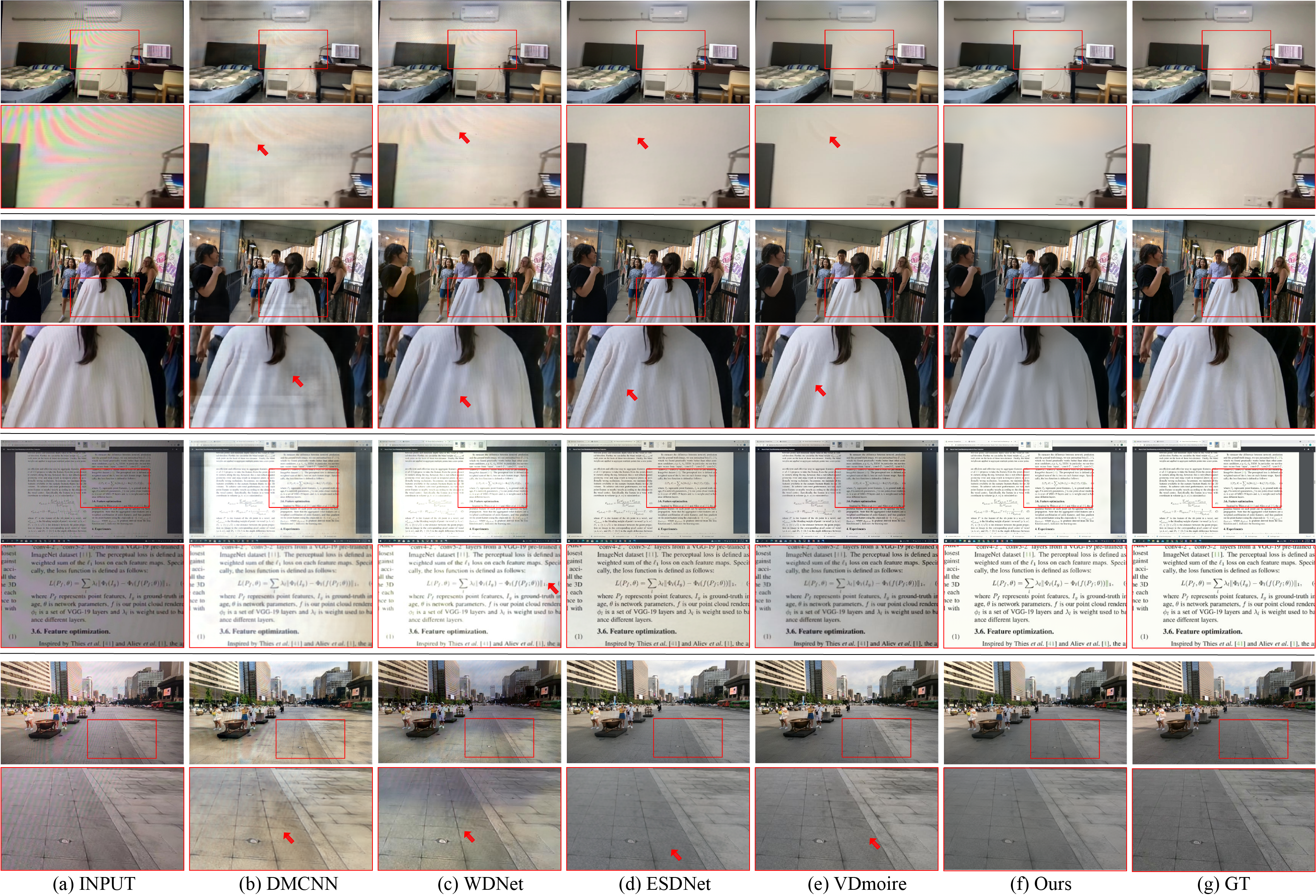}
\caption{Visual quality comparison among DMCNN, WDNet, ESDNet, VDMoire, and our proposed DTNet.}
\label{fig:visualconparison}
\end{figure*}

\subsubsection{Quantitative Results}
The performance of demoiréing is quantitatively measured using PSNR, SSIM, and LPIPS.
In Table~\ref{table:quantative-sota}, our proposed DTNet achieves leading video demoiréing performance on all four datasets. For instance, on the TCL-V1 dataset, we achieve a PSNR gain of 2.093 dB and an SSIM improvement of 0.067. Also, our approach significantly outperforms previous methods in terms of PSNR on both the TCL-V2 and iPhone-V2 datasets, exceeding 26 dB. Moreover, our method accomplishes a significant decrease in LPIPS, indicating a higher perceptual quality of the recovered images.

\begin{table}[htbp]
\caption{Quantitative comparison in terms of FVD and FSIM metrics to further analyze the quality of generated videos (or image frames) on TCL-V2 dataset. Note that lower FVD scores and higher FSIM scores indicate better performance.}
\label{table:video_comp}
\centering
        \begin{tabular}{lcc} 
            \toprule
            {Method}  & FVD$\downarrow$ & FSIM$\uparrow$ \\\midrule
            DMCNN   & 686.23 & 0.927 \\
            WDNet   & 713.52 & 0.936 \\
            ESDNet & 201.79 & 0.966 \\
            VDmoiré   & 269.13 & 0.960 \\
            Ours & \textbf{158.24} & \textbf{0.972} \\
            \bottomrule
        \end{tabular}
\end{table}

\subsubsection{Qualitative Results}
We present visual comparisons between our DTNet and the existing methods in Fig.~\ref{fig:visualconparison}. The results clearly demonstrate the advantages of our approach in removing moiré artifacts, particularly in the case of moiré patterns on blank walls or white T-shirts. Also, for scenes with rich details and textures, our method excels not only in correcting color shifts but also in restoring details.

\subsection{Video-Level Comparison}

Additionally, we employ two metrics, FVD ~\cite{unterthiner2018towards} and FSIM ~\cite{zhang2011fsim}, to assess the quality of video outputs. FVD adapts Frechet Inception Distance (FID) to evaluate the temporal coherence of a video, whereas FSIM emphasizes low-level features as an image quality assessment metric inspired by the human visual system. As demonstrated in Table~\ref{table:video_comp}, our model surpasses other existing approaches in both metrics, indicating that our outputs exhibit a greater similarity to the target distribution of the entire video sequences, while preserving per-pixel and structural visual information.

\subsection{Ablation Study}
Table~\ref{table:ablation} presents an assessment of the effectiveness of our proposed FDDA and TDR through ablation experiments involving diverse combinations of these foundational components. Furthermore, we investigate the necessity of multi-frame input by altering the input to repetitions of a single frame (Ours\_S). Note that, without loss of generality, we use the TCL-V2 dataset for the ablation study.

\begin{table}[!t]
\caption{Ablation Study of different variants of the proposed network structure.}
\label{table:ablation}
\centering
        \begin{tabular}{lccc} 
            \toprule
            Structure & PSNR$\uparrow$ & SSIM$\uparrow$ & LPIPS$\downarrow$  \\\midrule
            w/o Directional DCT & 26.005 & 0.872 & 0.140\\
            w/o Alignment & 23.635 & 0.848 & 0.196\\
            w/o Temporal-guided BL & 25.813 & 0.871 & 0.144\\
            Ours\_S & 25.914 & 0.870 & 0.145 \\
            Ours & \textbf{26.153} & \textbf{0.877} & \textbf{0.128}\\
            \bottomrule
        \end{tabular}
\end{table}

\begin{figure}[htbp]
\centering
\includegraphics[width=1\linewidth]{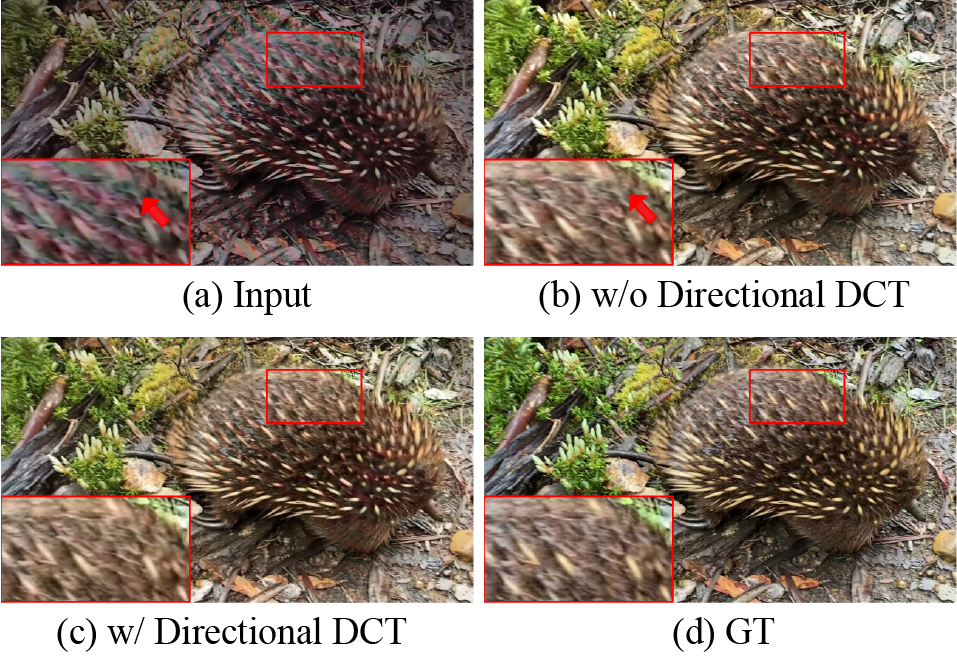}
\caption{Effect of Directional DCT.}
\label{fig:ab_demoire}
\end{figure}

\textbf{Directional DCT.}
Without applying directional DCT in the direction-aware demoiréing module, there is a decline in all three qualitative metrics. Upon observing the generated demoiréd images visually in complex scenarios, directional DCT demonstrates \textit{significant} contributions to moiré pattern removal.
Fig.~\ref{fig:ab_demoire} (a) displays an image of a hedgehog with alternating red and green moiré stripes. Due to the rich structural details, which consist of numerous edge features, the separation of moiré patterns from the original image content becomes highly challenging. When the directional DCT is not applied, Fig.~\ref{fig:ab_demoire} (b) exhibits noticeable residual moiré stripes on the hedgehog's back. In contrast, Fig.~\ref{fig:ab_demoire} (c) reveals a clear content which is less affected by moiré patterns, resembling the ground truth shown in Fig.~\ref{fig:ab_demoire} (d).

\begin{figure}[htbp]
\centering
\includegraphics[width=1\linewidth]{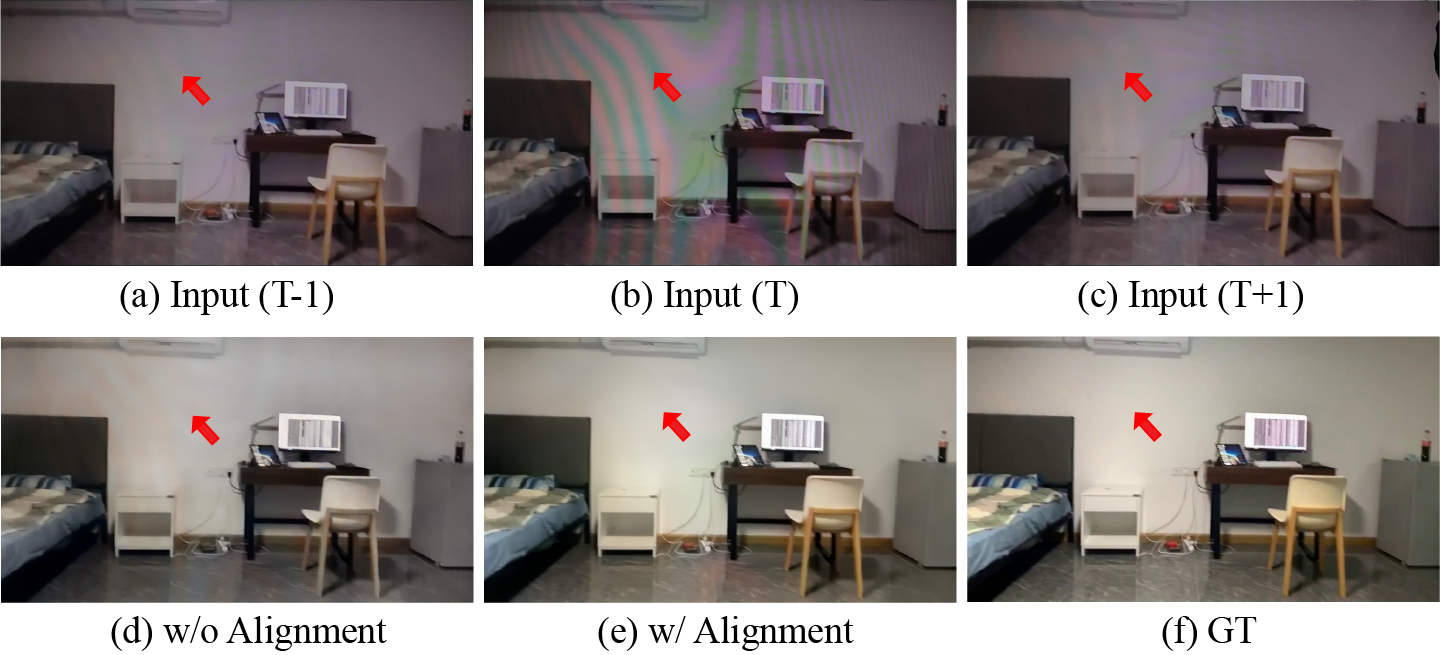}
\caption{Effect of Alignment.}
\label{fig:ab_alignment}
\end{figure}

\textbf{Alignment.}
In order to facilitate the utilization of information from neighboring frames and effectively handle significant and intricate motions, we employ both multi-scale alignment and deformable convolution for the alignment process. To demonstrate the indispensability of alignment process, we substitute the original design with a simple concatenation and two convolutional layers, resulting in a PSNR of only 23.635 dB. Fig.~\ref{fig:ab_alignment} (b) represents the frame we aim to restore, with extensive colorful moiré patterns on the blank wall in the center of the image. However, in the same location of the neighboring frames, as depicted in Fig.~\ref{fig:ab_alignment} (a, c), are less affected by moiré patterns. By utilizing a suitable alignment structure, we can effectively leverage the information from neighboring frames to assist in restoring the current frame, as demonstrated in Fig.~\ref{fig:ab_alignment} (e).

\begin{figure}[htbp]
\centering
\includegraphics[width=1\linewidth]{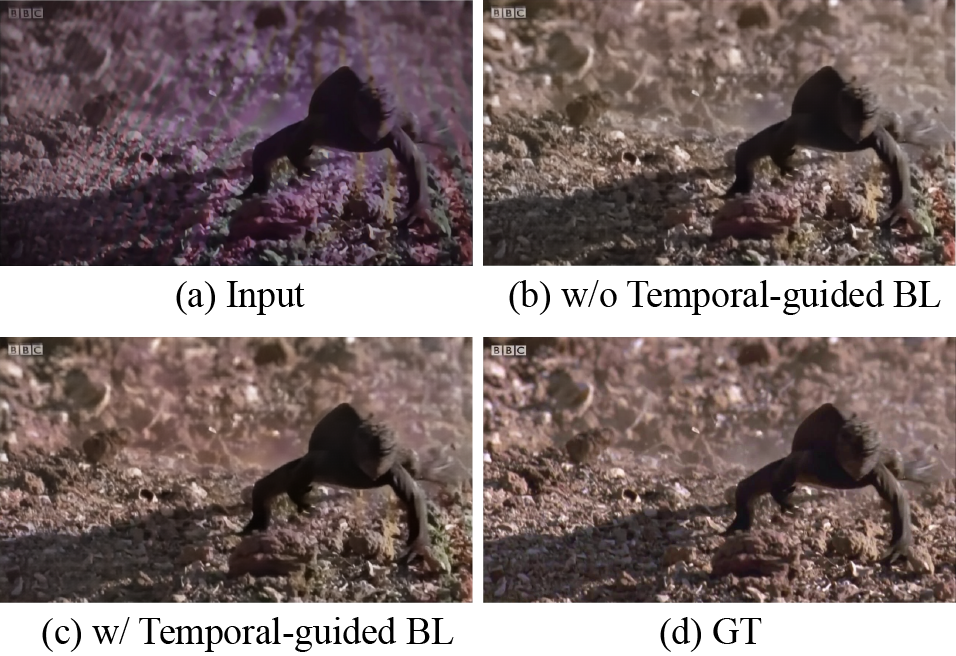}
\caption{Effect of Temporal-guided Bilateral Learning.}
\label{fig:ab_tdr}
\end{figure}

\textbf{Temporal-guided Bilateral Learning.}
In TDR, we propose the temporal-guided bilateral learning to address the color deviation and preserve image details. In the ablation study, we replace the proposed temporal-guided bilateral learning (Temporal-guided BL) with a concatenation followed by two convolutional layers for feature reconstruction, resulting in a PSNR drop of 0.35 dB. Also, the perceptual quality of the recovered images, as indicated by LPIPS scores, suffers a great decline. Arising from the ambient light and the screen itself, severe color degradation can be observed in Fig.~\ref{fig:ab_tdr} (a). With the absence of Temporal-guided BL, the color deviations remain challenging to rectify effectively. As a comparison, in Fig.~\ref{fig:ab_tdr} (c), the image generated using the Temporal-guided BL exhibit the correct tone, approaching the ground truth shown in Fig.~\ref{fig:ab_tdr} (d).

\section{Conclusion}
In this paper, we introduce DTNet, a unified framework for video demoiréing. In the FDDA process, we analyze the directional characteristics of moiré patterns and employ multiple directional DCT modes to perform the moiré pattern removal process in the frequency domain, thereby aiding the coarse-to-fine alignment process. In TDR, we learn localized tone curves guided by temporally aligned features to reduce color and detail loss caused by screen capture, while respecting the edges of latent clean images. Extensive frame-level and video-level experiments demonstrate that our video demoiréing method outperforms state-of-the-art approaches in both quantitative and qualitative evaluations.

\section{Acknowledgments}
This work was supported in part by Macau Science and Technology Development Fund under SKLIOTSC-2021-2023, 0072/2020/AMJ and 0022/2022/A1; in part by Research Committee at University of Macau under MYRG2022-00152-FST and  MYRG-GRG2023-00058-FST-UMDF; in part by Natural Science Foundation of China under 61971476; and in part by Alibaba Group through Alibaba Innovative Research Program.

\bibliography{DTNet}

\newpage
\clearpage

\section{Appendix}

\section{Model Complexity Comparison}
In Table~\ref{tab:model_complexity}, the numbers of parameters (Params) and multiply-accumulation operations (MACs) are used to compare the model complexity. Also, we provide the corresponding quantitative results of the image demoiréing methods, i.e., DMCNN \cite{sun2018moire}, WDNet \cite{liu2020wavelet}, ESDNet \cite{yu2022towards} and the video demoiréing method VDmoiré \cite{dai2022video} on the iPhone-V1 dataset.

\begin{table}[htbp]
\setlength\tabcolsep{5.5pt}
\centering
\caption{Quantitative and the model complexity comparison on VDmoiré dataset (iPhone-V1). The best results are in \textbf{bold} and the second best results are \underline{underlined}.}
\label{tab:model_complexity}
\scalebox{0.83}{
    \begin{tabular}{lccc|cc}
    \toprule[1pt]
    Method & PSNR$\uparrow$ & SSIM$\uparrow$ & LPIPS$\downarrow$ & Params(M)& MACs(T)  \\
    \midrule[1pt]
    DMCNN  & 21.967 & 0.712  & 0.280 & \textbf{1.43}  & 0.368 \\
    WDNet & 19.818& 0.722 & 0.300 & 3.92  & 1.550\\
    ESDNet& 22.537 & 0.731 & 0.218 & 5.93  & 2.242 \\
    VDmoiré & 21.990 & 0.707 & 0.221 & 5.98 & 0.608 \\
    \midrule[0.5pt]
    DTNet-light & \underline{24.473} & \underline{0.782} & \underline{0.173} & \underline{3.69} & \textbf{0.233} \\
    DTNet-full & \textbf{24.821} & \textbf{0.794} & \textbf{0.172} & 7.36 & \underline{0.587} \\
    \bottomrule[1pt]
    \end{tabular}}
\end{table}

Our proposed DTNet-full outperforms the second best method ESDNet by a large margin, \textit{i}.\textit{e}., 2.284 dB in PSNR, 0.063 in SSIM, and 0.046 in LPIPS. For the model complexity, DTNet-light achieves the second-lowest numbers of parameters with 3.69M and the lowest computation overhead with 0.233T MACs. Meanwhile, DTNet-light still maintains a performance gain of 1.936 dB in PSNR over ESDNet.

\section{More Frame-level Visualization Comparison}
\begin{figure*}[htbp]
\centering
\includegraphics[width=1\linewidth]{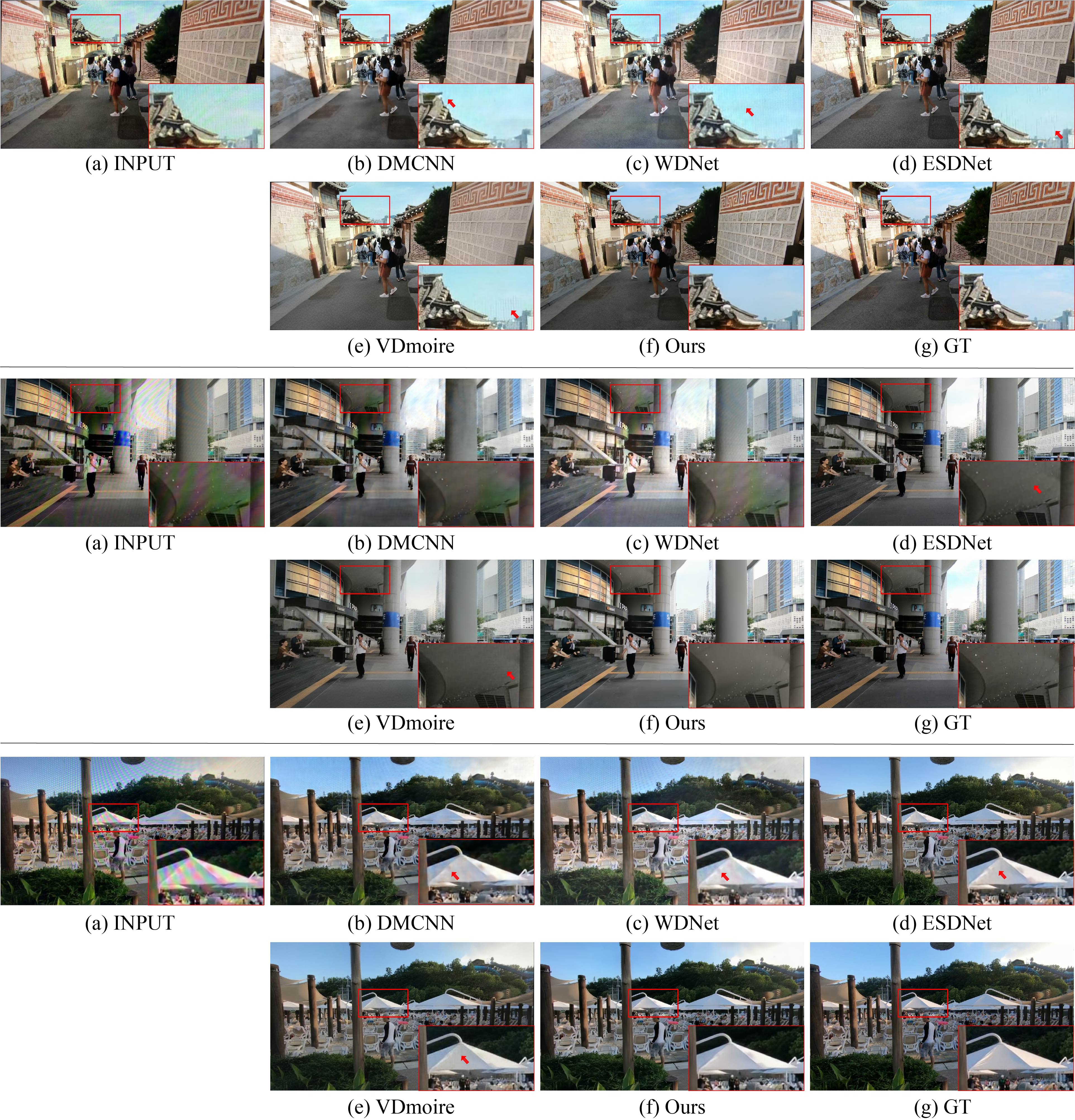}
\caption{Visual quality comparison among DMCNN, WDNet, ESDNet, VDmoiré, and our proposed DTNet.}
\label{fig:suppl_Q_3}
\end{figure*}

\begin{figure*}[htbp]
\centering
\includegraphics[width=1\linewidth]{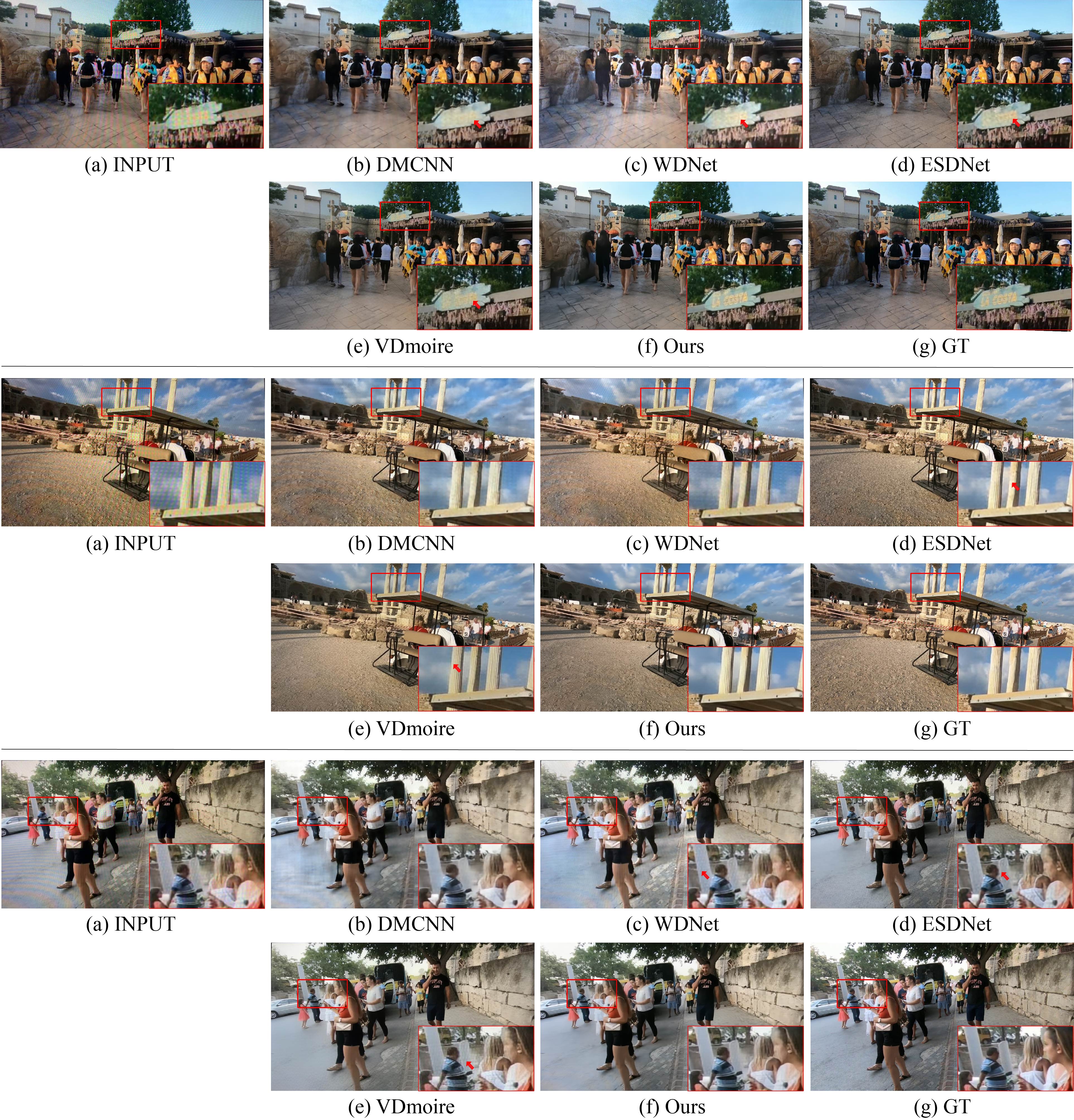}
\caption{Visual quality comparison among DMCNN, WDNet, ESDNet, VDmoiré, and our proposed DTNet.}
\label{fig:suppl_Q_1}
\end{figure*}

\begin{figure*}[htbp]
\centering
\includegraphics[width=1\linewidth]{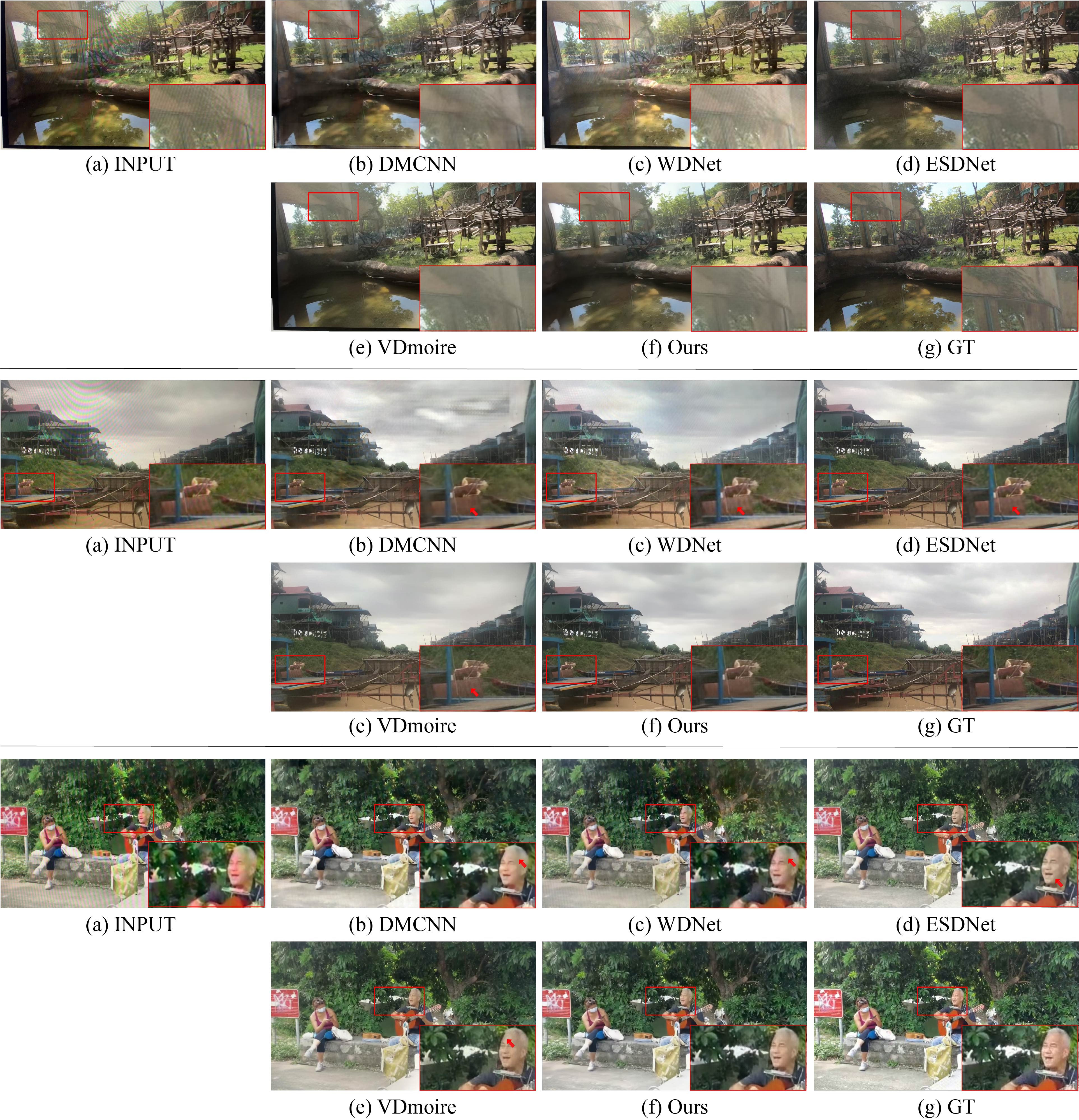}
\caption{Visual quality comparison among DMCNN, WDNet, ESDNet, VDmoiré, and our proposed DTNet.}
\label{fig:suppl_Q_2}
\end{figure*}

\begin{figure*}[htbp]
\centering
\includegraphics[width=1\linewidth]{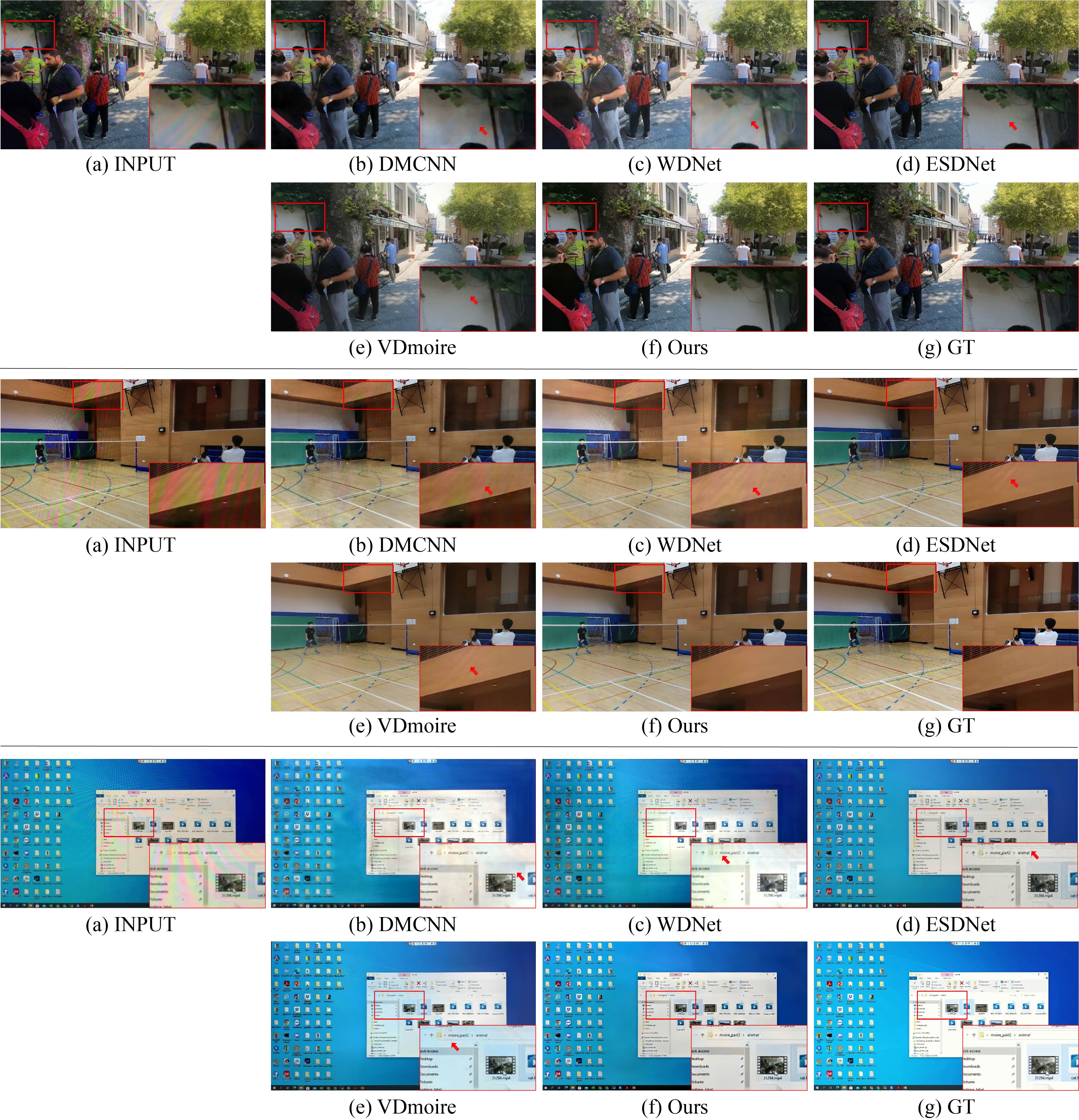}
\caption{Visual quality comparison among DMCNN, WDNet, ESDNet, VDmoiré, and our proposed DTNet.}
\label{fig:suppl_Q_4}
\end{figure*}

We provide more visual comparisons with state-of-the-arts methods in Figs. \ref{fig:suppl_Q_3}-\ref{fig:suppl_Q_4}. Some regions are zoomed for better observation and comparison. The results clearly demonstrate the advantages of our proposed DTNet in removing moiré patterns. Also, for scenes with rich details and textures, our method excels not only in correcting color shifts but also in restoring details.

For instance, in the first scene illustrated in Fig. \ref{fig:suppl_Q_3}, images generated by all the previous methods exhibit color deviations when compared to the ground truth image. Furthermore, upon detailed scrutiny of the magnified area, artifacts resembling vertical stripes are evident in the sky background of images produced by other methods. In contrast, our approach effectively eliminates moiré patterns and rectifies color deviations.
In the first scene depicted in Fig. \ref{fig:suppl_Q_1}, the textual information on a distant sign is contaminated by moiré patterns. In images generated by previous methods, the moiré patterns on the sign have not been sufficiently eradicated, resulting in unclear textual information. In comparison, our method is capable of restoring rich details and textures, delivering a superior visual experience. Similar observations can also be obtained in Figs. \ref{fig:suppl_Q_2}-\ref{fig:suppl_Q_4}.

\newpage

\section{Video-level Visualization Comparison}
To evaluate the performance of the generated videos, we provide the visualized comparison on the video quality (video link: \textit{https://youtu.be/mBCxKpy8Cig}). The frame rate of the videos is set to 10fps.

As can be seen from the video, our proposed DTNet can not only accommplish frame-level moiré pattern removal and color correction, but also achieve the temporal consistency.
Both DMCNN and WDNet fall short in achieving satisfactory moiré pattern removal. Notably, prominent moiré patterns persist on the blank wall in Scene 1 and the white T-shirt in Scene 2. Furthermore, in Scene 3, both methods introduce severe color deviations to the overall image, compromising the viewer's visual experience.
In terms of temporal consistency, in Scene 1, previous image demoiréing methods lead to abrupt visual transitions between consecutive frames. Even the former video demoiréing method, VDmoiré, fails to achieve optimal temporal consistency. Although VDmoiré introduces an alignment process, it directly aligns the features from original frames containing moiré patterns,thereby influencing the alignment performance to a certain degree. These issues are effectively addressed in our DTNet, where we employ a direction-aware demoiréing module to efficiently eliminate moire patterns. Additionally, conducting this pre-demoiré operation before alignment facilitates the extraction of information from adjacent frames, enabling effective management of intricate motions. Therefore, temporal consistency is ensured in our DTNet.

\end{document}